\documentclass[runningheads]{llncs}

\usepackage[T1]{fontenc}
\usepackage{graphicx}
\usepackage{booktabs}
\usepackage{soul}
\usepackage{comment}

\soulregister{\cite}{7}
\soulregister{\ref}{7}
\soulregister{\pageref}{7}

\usepackage[font=small,labelfont=bf,up,textfont=it,up]{caption}
\usepackage{subfig}
\usepackage{float}
\usepackage{tabularx}
\usepackage{multirow}

\usepackage{longtable}
\usepackage{multirow}
\usepackage{rotating}

\begin{document}
\title{UlcerGPT: A Multimodal Approach Leveraging Large Language and Vision Models for Diabetic Foot Ulcer Image Transcription}
%
\titlerunning{UlcerGPT}
%
\author{Reza Basiri\inst{1,2}\orcidID{0000-0002-0209-6478} \and
Ali Abedi\inst{1,2} \and
Chau Nguyen\inst{1,3} \and
Milos R. Popovic\inst{1,2} \and
Shehroz S. Khan\inst{1,2}}
\authorrunning{Basiri et al.}
%
\institute{KITE Research Institute, University Health Network, Toronto, Canada \and
Institute of Biomedical Engineering, University of Toronto, Toronto, Canada \and
Mathematical and Computational Sciences, University of Toronto Mississauga, Canada
\\
\email
{
\{reza.basiri,chauminh.nguyen\}@mail.utoronto.ca
\{ali.abedi,milos.popovic,shehroz.khan\}@uhn.ca
}
}
\maketitle 
\begin{abstract}
Diabetic foot ulcers (DFUs) are a leading cause of hospitalizations and lower limb amputations, placing a substantial burden on patients and healthcare systems. Early detection and accurate classification of DFUs are critical for preventing serious complications, yet many patients experience delays in receiving care due to limited access to specialized services. Telehealth has emerged as a promising solution, improving access to care and reducing the need for in-person visits. The integration of artificial intelligence and pattern recognition into telemedicine has further enhanced DFU management by enabling automatic detection, classification, and monitoring from images. Despite advancements in artificial intelligence-driven approaches for DFU image analysis, the application of large language models for DFU image transcription has not yet been explored. To address this gap, we introduce UlcerGPT, a novel multimodal approach leveraging large language and vision models for DFU image transcription. This framework combines advanced vision and language models, such as Large Language and Vision Assistant and Chat Generative Pre-trained Transformer, to transcribe DFU images by jointly detecting, classifying, and localizing regions of interest. Through detailed experiments on a public dataset, evaluated by expert clinicians, UlcerGPT demonstrates promising results in the accuracy and efficiency of DFU transcription, offering potential support for clinicians in delivering timely care via telemedicine.

\end{abstract}

\section{Introduction}
Diabetes is a rapidly growing global health issue, with more than 537 million adults affected worldwide as of 2021. There is a projection that this number will reach 783 million by 2045, reflecting the increasing prevalence of the disease across various populations \cite{ong2023global}. Among the many complications associated with diabetes, Diabetic Foot Ulcers (DFUs) \cite{armstrong2023diabetic} are among the most severe and challenging to manage. Affecting approximately $15-25\%$ of diabetic patients during their lifetime, DFUs are a leading cause of hospitalizations and lower limb amputations \cite{manji2024effectiveness,basiri2021reduction,armstrong2023diabetic}. The burden of DFUs on patients' quality of life and healthcare systems underscores the critical need for effective management strategies \cite{dardari2023hospital}.\\

Research indicates that early intervention can significantly reduce the risk of lower extremity amputations by a high proportion \cite{armstrong2023diabetic}. However, many patients face delays in receiving appropriate care due to limited access to specialized healthcare services and geographic barriers \cite{foong2020facilitators}. Integrating Artificial intelligence (AI) and pattern recognition into telemedicine enhances DFU management by enabling automatic detection, classification, and monitoring from images \cite{shiraishi2024appropriateness}. These tools offer accurate, quick assessments, support clinicians in prioritizing urgent cases, and ensure consistent monitoring in telehealth settings.\\

Large Language Models (LLMs) have recently gained significant attention due to their capability to process and understand text at an advanced level. Models such as Generative Pre-trained Transformers (GPT) \cite{radford2018improving} have become widely used across various domains, including healthcare, for tasks such as text generation, question answering, and managing complex information \cite{thirunavukarasu2023large}. In image analysis, Vision Transformers (ViTs) \cite{fang2023eva} have made substantial advancements. ViTs can process visual data, such as medical images, and convert it into a format that LLMs can utilize, facilitating the integration of image and text processing. This capability enhances the interpretation and analysis of medical images \cite{nerella2024transformers}.\\

A multitude of AI-driven approaches have been proposed for the identification, detection, classification, and localization of DFUs from foot images, utilizing techniques from machine learning, deep learning, and computer vision \cite{basiri2024protocol,zhang2022comprehensive,basiri2023synthesizing}. However, to the best of our knowledge, the application of LLMs for DFU image transcription remains unexplored. DFU image transcription is essential for facilitating telemedicine by enabling timely detection and identification of ulcers, which supports clinicians in providing prompt and effective care. To address the gap in this field, this paper introduces UlcerGPT, a novel multimodal approach that leverages large language and vision models to transcribe and analyze DFU images. This method aims to enhance the accuracy and efficiency of DFU detection and classification, thereby further supporting clinicians in telemedicine. This work makes the following contributions:

\begin{itemize}    
    \item By combining advanced vision and language models, such as Large Language and Vision Assistant (LLaVA) \cite{liu2023llava} and Chat Generative Pre-trained Transformer (ChatGPT) \cite{openai2024chatgpt}, a new deep-learning framework was introduced to transcribe DFU images by jointly detecting, tokenizing, and narrating DFU’s elements of interest.
    \item Detailed experiments were conducted on a public dataset \cite{yap2022diabetic}, with the transcription results of the proposed method evaluated by expert clinicians, demonstrating its effectiveness in DFU image transcription.
\end{itemize}

The structure of this paper is organized as follows. Section \ref{sec:related_work} offers an overview of the relevant literature. This is succeeded by Section \ref{sec:methodology}, which details the proposed methodology. Following this, Section \ref{sec:experiments} describes the experimental setup and discusses the results obtained with the proposed method. Lastly, Section \ref{sec:conclusion} concludes the paper and proposes directions for future research.

\section{Related Work}
\label{sec:related_work}
This section begins with a review of recent AI-driven approaches for DFU image analysis, followed by an examination of prior works that have integrated large language and vision models for the analysis of medical images.

\subsection{AI-driven DFU Image Analysis}
\label{sec:related_work_dfu}
Zhang et al. \cite{zhang2022comprehensive} conducted a literature review on deep-learning approaches for the classification, object detection, and semantic segmentation of DFU images. Zhang et al. identified that, for classification tasks in DFU imaging, the most effective models were all based on Convolutional Neural Networks (CNNs). For object detection tasks, the leading models utilized architectures such as Faster R-CNN \cite{da2021faster}, and EfficientDet \cite{goyal2020refined}. In semantic segmentation tasks, models based on fully convolutional networks (FCNs), U-Net, V-Net, and SegNet were employed, with U-Net achieving the highest accuracy at 94.96\% \cite{rania2020semantic}. The most recent methods for DFU image analysis have predominantly employed vision transformers, reflecting a shift towards leveraging advanced transformer-based architectures for improved performance in this domain \cite{10242064,brodzicki2023dfu}.

\subsection{Large Language and Vision Models for Medical Image Analysis}
\label{sec:related_work_llm}
Given the absence of prior approaches integrating LLMs and vision models for DFU image analysis, this subsection reviews previous applications in analyzing medical images in other domains. Hu et al. \cite{hu2024advancing} conducted a literature review on the application of LLMs in medical images, covering applications such as image captioning, report generation, and visual question-answering across various domains such as MRI, CT, ultrasound, and chest X-rays. The following discussion focuses on the latest methods involving LLMs and vision models, particularly in chest X-ray imaging, as representative examples of medical imaging applications.\\

Wiehe et al. \cite{wiehe2022language} explored the adaptation of CLIP-based models \cite{radford2021learning} for classifying chest X-ray images. Since the features learned by pre-trained CLIP models on general internet data do not directly transfer to the chest X-ray domain, the authors adapted CLIP to chest radiography using contrastive language supervision. This adaptation resulted in a model that outperformed supervised learning approaches on the MIMIC-CXR dataset \cite{johnson2019mimic}. Additionally, language supervision improved model explainability, enabling the multi-modal model to generate images from text, allowing experts to inspect what the model has learned.\\

Thawkar et al. \cite{thawakar2024xraygpt} introduced XrayGPT, a conversational medical vision-language model tailored for analyzing and responding to open-ended questions related to chest X-rays. This model is based on the latest advancements in LLMs, such as Bard and GPT-4, while addressing the specific challenges associated with interpreting biomedical images in the radiology domain. XrayGPT achieves this by aligning a medical visual encoder (MedClip) \cite{wang2022medclip} with a fine-tuned LLM (Vicuna) \cite{chiang2023vicuna} via a linear transformation, allowing the model to excel in visual conversation tasks that require medical expertise. Additionally, to enhance the model's performance, the authors generated a large set of interactive and high-quality summaries from free-text radiology reports, which were employed for fine-tuning the LLM.\\

Inspired by the related works in the other medical domains, the next section introduces UlcerGPT, a novel multimodal approach for transcription and analysis of DFU images, to address the gap in the use of large language and vision models for DFU image analysis.

\section{Method}
\label{sec:methodology}
The study utilizes a dataset from the DFU2022 competition \cite{yap2024diabetic}, which provides a comprehensive collection of 2,000 annotated clinical RGB images specifically related to DFUs. The dataset was initially developed for research in detecting and classifying DFUs. The images mainly include the plantar aspect of the foot with one or more ulcerations and background drapes, minimalizing the presence of other elements in the images. These images are selected to represent a variety of DFU cases, including different stages of ulceration, anatomical locations on the foot, and associated skin conditions, ensuring a robust test set for evaluating the LLMs. Additionally, this dataset is only accessible for research and not hosted on public domains, so the available LLMs have not been previously trained on this dataset.\\

\begin{figure}[ht!]
\centering
\includegraphics[width=0.9\textwidth]{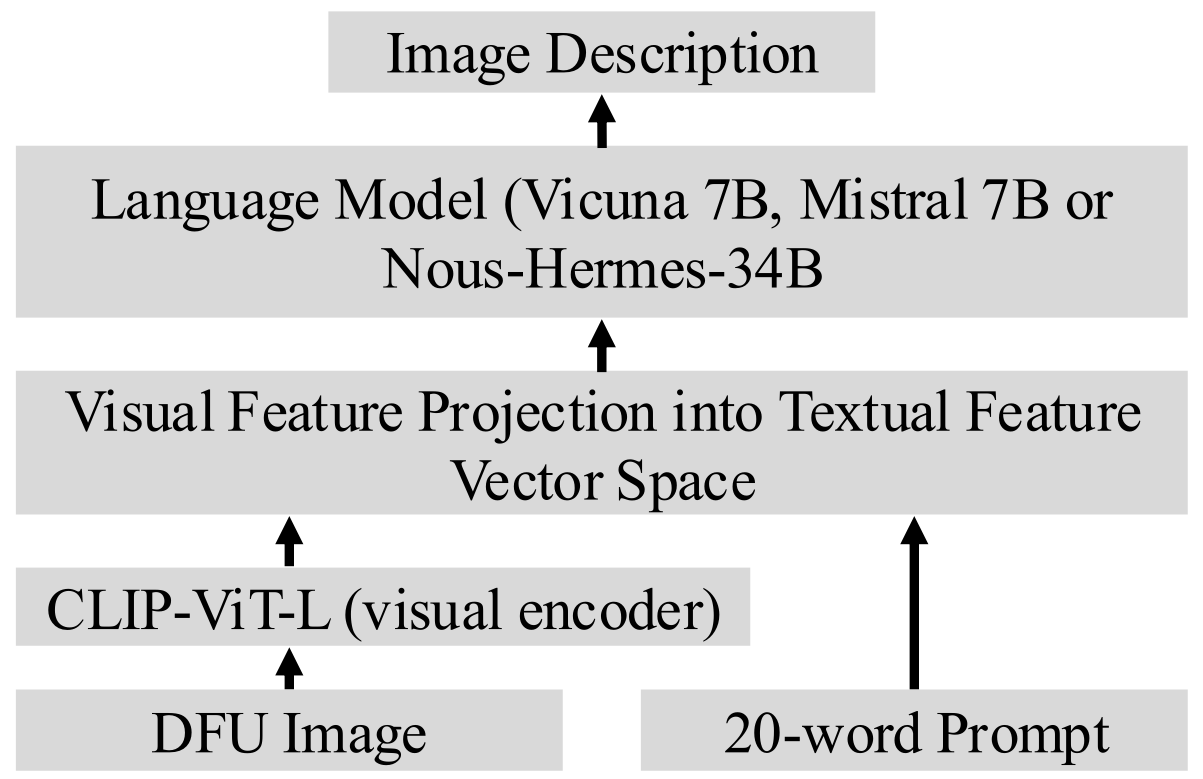}
\caption{Overview of the LLaVA setup used in the study, illustrating the constant backbone architecture with varying language models to assess their influence on performance independent of the vision-language components.} \label{fig3}
\end{figure}

The models evaluated in this study include GPT-4omni (GPT-4o) \cite{openai2024}, Qwen-VL \cite{bai2023qwen}, LLaVA integrated with Nous-Hermes \cite{nousresearch2024noushermes}, including 34B-parameter, LLaVA combined with Mistral \cite{jiang2023mistral} with a 7B-parameter LLM backbone, and LLaVA paired with Vicuna \cite{vicuna2023} with a 7B-parameter LLM backbone. For GPT4o, the gpt-4o-2024-08-06 snapshot was used. In the LLaVA setup, as shown in Figure \ref{fig3}, the CLIP tokenizer backbone was kept constant while the language model part of the architecture varied to investigate the language model influences independent from CLIP and other components of a vision-language architecture. \\

These models were chosen for their state-of-the-art performance in language processing, with designs to handle both visual and textual data, making them particularly relevant for generating accurate and clinically relevant descriptions of DFU images. Additionally, the models were selected to include commercial and open-source and different parameter sizes to comprehensively evaluate their applications in DFU. All models except GPT-4o are open-sourced and deployed on a Nvidia 32GB machine by loading the relevant weights from the official GitHub or Hugging Face repositories. For GPT-4o, the OpenAI website platform was used.\\

Each model was tasked with generating a brief, clinically-focused description of the DFU images. The following prompt was used: 
\begin{center}
"In about 20 words, describe this image to a medical doctor. The doctor may use the description to complete the EMR." 
\end{center}

The prompt used was standardized to ensure consistency across models, asking each to describe the image in a manner that a medical doctor might use to complete an electronic medical record (EMR). The prompt specified a 20-word limit to ensure the description included terms or synonyms for “Diabetic,” “Foot,” “Ulcer,” “Plantar,” “Amputation,” and “Calluses.” Each DFU and amputation location should be described with 3 to 4 words (e.g., "one sub-second metatarsal") along with one preposition or article for each noun, totalling 20 words. The models' performances were evaluated based on several critical metrics: clinical accuracy, comprehensiveness, location accuracy, and diagnostic utility. Clinical accuracy measures the fidelity of the description to the visual and clinical details of the DFU. Comprehensiveness assesses how thoroughly the description covers relevant aspects of the ulcer, while location accuracy evaluates the precision with which the model identifies and describes the ulcer's position on the foot. Diagnostic utility reflects the usefulness of the description in supporting clinical decision-making. A description with high diagnostic utility would provide information that helps form a diagnosis, guide treatment, or monitor the patient's progress. For example, if a description includes relevant details about infection, tissue necrosis, or healing stages, it would score higher in this category. The clinician evaluators were provided with a brief description of each metric.\\

The evaluation involved a panel of 2 clinicians with expertise in diabetic foot care from the Zivot Limb Preservation Centre, Calgary, Canada, who independently evaluated the model-generated descriptions of five DFU images. Each clinician received the images and corresponding generated descriptions via digital media. Each image, along with the generated text descriptions, Likert scale, and a table defining the evaluation criteria (clinical accuracy, comprehensiveness, location accuracy, and diagnostic utility), was presented on a separate page. The clinicians individually assessed the descriptions for each image using a 5-point Likert scale (1 = Poor, 5 = Excellent), without consulting each other. Their ratings were then averaged to determine the overall performance of each model. The five DFU images were selected to provide a diverse range of possible DFU conditions. The selected image had one of the following characteristics: 1. Simple undebrided DFU with hyperkeratosis, 2. Multiple DFUs in one image, 3. Presence of gangrene 4. Amputated toes, and 5. Post-debridement DFU with visible granulation. The clinicians’ ratings were averaged to produce an overall assessment of each model’s performance. The average values were plotted on a spider plot. Statistical analyses, including the calculation of means, standard deviations, and ANOVA, were conducted to identify any significant differences in model performance.

\section{Experiments}
\label{sec:experiments}

\begin{figure}[h!]
    \centering
    \subfloat[]{\includegraphics[width=0.3\textwidth,height=3.5cm]{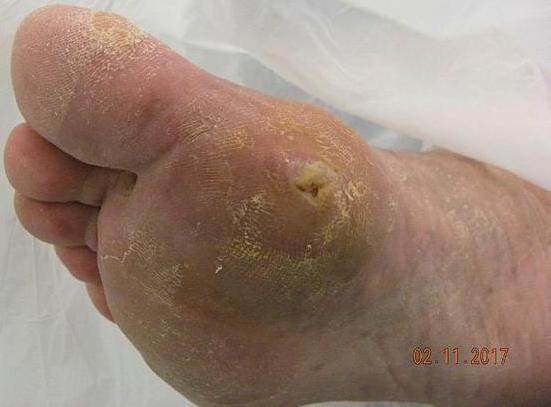}}
    \hfill
    \subfloat[]{\includegraphics[width=0.3\textwidth,height=3.5cm]{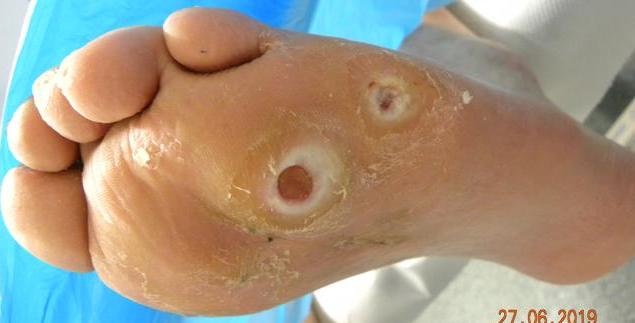}}
    \hfill
    \subfloat[]{\includegraphics[width=0.3\textwidth,height=3.5cm]{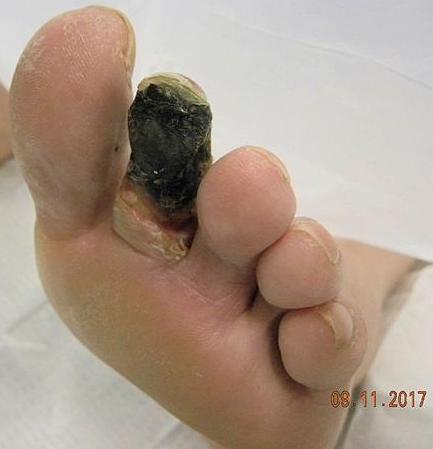}}
    \vspace{0.5cm}
    \subfloat[]{\includegraphics[width=0.3\textwidth,height=3.5cm]{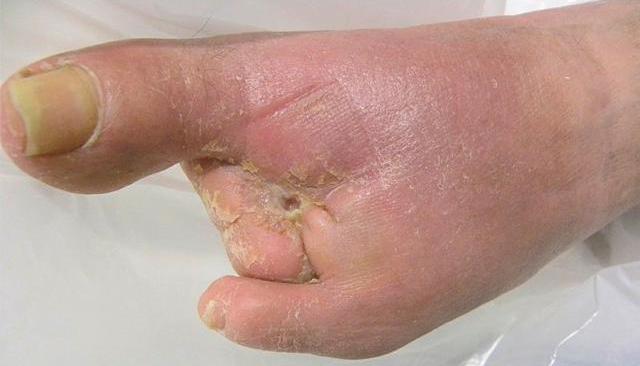}}
    \hspace{0.1\textwidth}
    \subfloat[]{\includegraphics[width=0.3\textwidth,height=3.5cm]{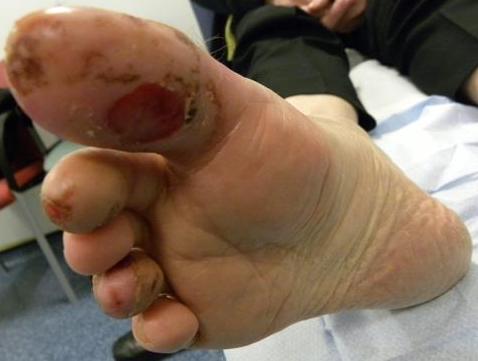}}
    \caption{Selected DFU images used in the study to evaluate model-generated descriptions. Five images (a-e) each containing unique clinical features were individually processed by LLMs and results were evaluated by the expert clinicians}
    \label{fig1}
\end{figure}

The five DFU images shown in Figure \ref{fig1} were selected to follow the criteria described in the methods. Descriptions generated from each of the models for Figure \ref{fig1} images are shown in Table 1. \\

\newpage
\begin{longtable}{|p{1.2cm}|p{\dimexpr\textwidth-2\tabcolsep-1cm\relax}|}
\caption{Baseline model outputs describing Figure 1 foot ulcers.}\label{tab1} \\

\hline
\textbf{Model} & \textbf{Baseline Output} \\ \hline
\endfirsthead

\multicolumn{2}{c}{Table \thetable{} continued from previous page} \\
\hline
\textbf{Model} & \textbf{Baseline Output} \\ \hline
\endhead

\hline \multicolumn{2}{r}{Continued on next page} \\
\endfoot

\hline
\endlastfoot

\multirow{5}[-4]{*}{\rotatebox[origin=c]{90}{GPT-4o}} & 
(a) Foot ulcer on the plantar surface near the first metatarsal, surrounded by dry, calloused skin with hyperkeratosis. \\
& (b) Multiple, deep, circular ulcers with hyperkeratotic borders on the plantar surface of a foot, likely suggesting diabetic foot ulcers or pressure sores. \\ 
& (c) Necrosis of the big toe, characterized by blackened tissue and signs of infection. The skin appears discolored with significant tissue damage, indicating gangrene. \\
& (d) A swollen foot with thickened skin, peeling, and ulceration between the toes, possibly indicating a fungal infection or severe dermatitis. \\
& (e) A foot with multiple ulcers and necrotic tissue, particularly affecting the toes, indicating possible severe diabetic foot complications or another ischemic condition. \\
\hline
\newpage
\multirow{5}[-3]{*}{\rotatebox[origin=c]{90}{Qwen-VL}} & 
(a) Patient's foot with a large, yellowish ulceration on the sole, surrounded by dry, cracked skin. \\ 
& (b) Two open wounds on its sole, one larger and central, the other smaller and uppermost, both surrounded by red, inflamed skin. \\
& (c) A blackened toe with necrotic tissue is seen on a foot, likely due to peripheral artery disease or diabetes. \\
& (d) A foot displays an open wound on its sole, surrounded by red, inflamed skin. \\
& (e) A foot displays a large, open wound on its big toe, surrounded by red, inflamed skin. \\
\hline

\multirow{5}[15]{*}{\rotatebox[origin=c]{90}{LLava + Nous-Hermes}} & 
(a) A foot with a callus on the heel and surrounding skin appears dry and flaky. \\ 
& (b) A foot with a central, circular, ulcerated lesion surrounded by a raised, erythematous border, indicative of a possible skin condition or infection. \\
& (c) Foot with a large, black, necrotic ulceration on the sole, surrounded by erythema and possibly some exudate. \\
& (d) A human foot with a wound on the heel area. The skin appears red and inflamed, with a small puncture in the center. \\
& (e) Patient's left foot with a red, inflamed, and possibly infected toenail. \\
\hline

\multirow{5}[10]{*}{\rotatebox[origin=c]{90}{LLaVA + Mistral}} & 
(a) A foot with skin lesions and infection on the big toe, indicating possible fungal or bacterial infection. \\ 
& (b) Foot with various sores and an infection visible on the bottom, showing signs of pus and inflammation, which might suggest bacterial or fungal infection. \\
& (c) A foot with an infected toe, with signs of inflammation. The surrounding skin is red and swollen, indicating an infection in progress. \\
& (d) The foot displays an infected toe with inflamed skin and visible pus near the nail bed, and yellowish discolorations on the toes. \\
& (e) A foot with a sore or injury on the toe. There are also red and white marks on the toes, which could indicate minor infections or bruising. \\
\hline
\multirow{5}[10]{*}{\rotatebox[origin=c]{90}{LLaVA + Vicuna}} & 
(a) A foot with an infected toe, likely bacterial or fungal, needing prompt medical attention to avoid complications. \\ 
& (b) The foot exhibits signs of a fungal infection, including discoloration, flaking skin, and a potential bacterial or viral lesion. \\
& (c) A foot with a severely cracked and infected toenail. The condition suggests poor hygiene, which may have led to infection. \\
& (d) The foot is lifted with a worn toenail, potentially indicating stress. It may be "ungual," leading to discomfort and infection if ignored. \\
& (e) A foot with what appears to be a toenail infection or fungus. There are visible signs of discoloration. \\
\hline

\end{longtable}

\begin{figure}[t!]
\centering
\includegraphics[width=0.8\textwidth]{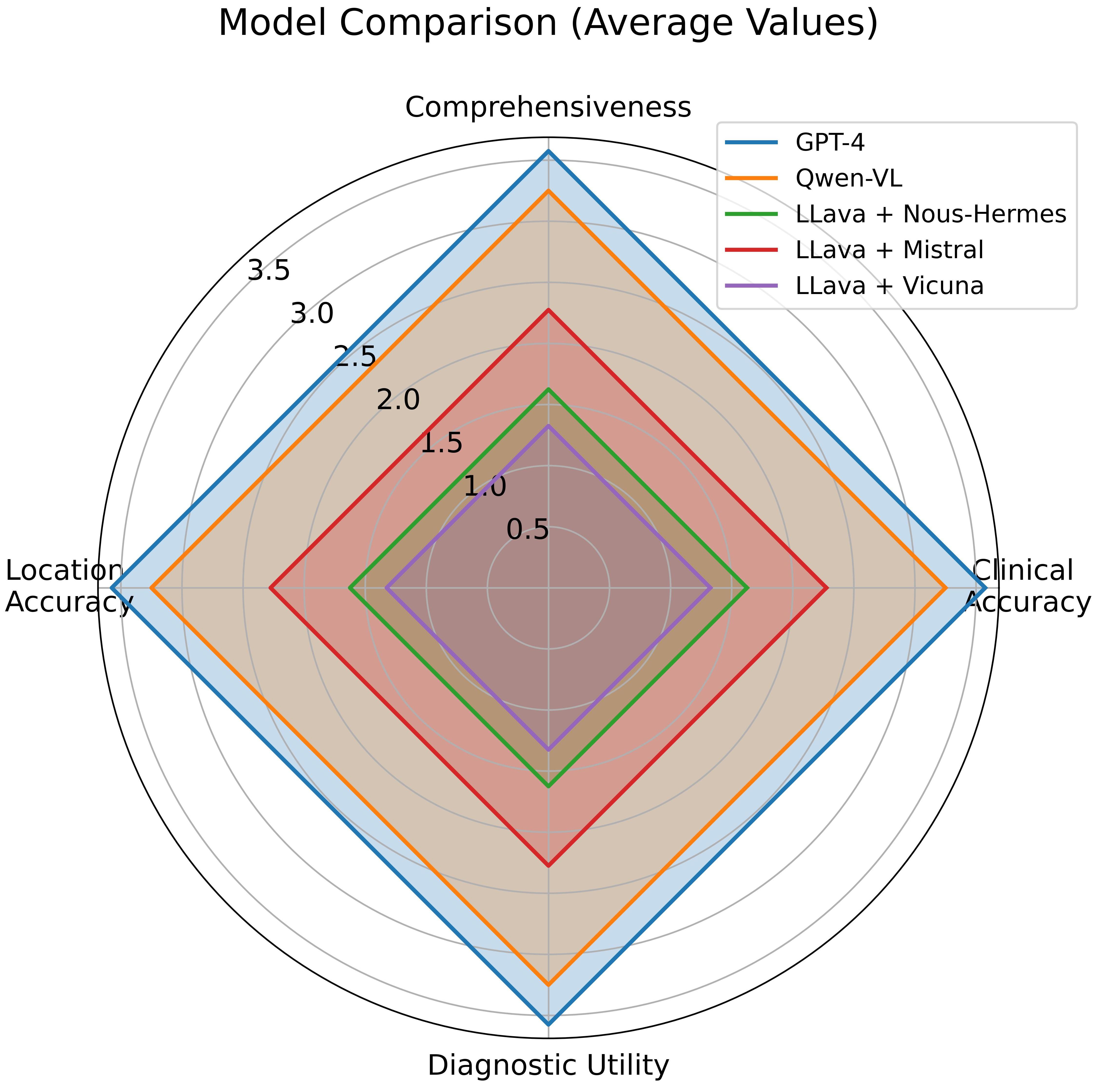}
\caption{Average performance comparison of language models across key DFU clinical-relevant metrics.} \label{fig2}
\end{figure}

Among the models evaluated, GPT-4o demonstrated the highest overall performance, with an overall average of 3.6 as shown in Figure \ref{fig2}. Figure \ref{fig2} illustrates the average performances of each model is a spider plot for easier comparison. Additionally, Table 1 includes each category's clinical evaluation breakdown separately. The results indicate that GPT-4o provides a reliable and consistent description of DFUs, accurately capturing the key clinical features and relevant details. Qwen-VL followed closely and was the highest-performing open-source model with an overall score of 3.3. While still effective, Qwen-VL's performance suggests it may miss some nuances in DFU descriptions that GPT-4 captures. LLaVA combined models performed significantly lower, with overall average scores of 2.3, 1.6 and 1.3 for Mistral, Nous-Hermes, and Vicuna variants, respectively. These results suggest that these versions of LLaVA struggle with accurately capturing the clinical context and relevant details needed for effective DFU descriptions. 

\begin{table}[ht!]
\caption{Clinicians' ratings of model-generated descriptions. Ratings are based on a Likert scale (1 to 5, where 1 = Poor, 5 = Excellent).}\label{tab2}
\centering
\begin{tabular}{|l|c|c|c|c|}
\hline
\textbf{Models} & \textbf{Clinical Accu.} & \textbf{Compre.} & \textbf{Location Accu.} & \textbf{Diagnostic Util.} \\ \hline
GPT-4o                 & 3.6 & 3.5 & 3.6 & 3.6  \\ \hline
Qwen-VL               & 3.3 & 3.3 & 3.3 & 3.1   \\ \hline
LLaVA + Nous-Hermes   & 1.6 & 1.7 & 1.6 & 1.6   \\ \hline
LLaVA + Mistral       & 2.3 & 2.3 & 2.3 & 2.2   \\ \hline
LLaVA + Vicuna        & 1.3 & 1.1 & 1.6 & 1.3   \\ \hline
        \multicolumn{5}{|c|}{Footnote: Accu. = Accuracy, Compre. = Comprehensiveness, Util. = Utility }\\ \hline
\end{tabular}
\end{table}
Statistical analysis using ANOVA confirmed that the differences observed in the performance metrics across the models were statistically significant, particularly in the comprehensiveness of the descriptions. The p-values obtained from the ANOVA tests were below the 0.05 threshold, indicating that the variations in performance between the models are unlikely due to random chance.

Inter-rater reliability between the two clinicians was measured using Cohen’s Kappa. The Kappa values for clinical accuracy, comprehensiveness, location accuracy, and diagnostic utility were 0.05, 0.10, 0.05, and 0.15, respectively, indicating slight agreement between the two evaluators. These low levels of agreement suggest that subjective differences in interpreting the generated descriptions may exist.

\newpage
\section{Conclusion}
\label{sec:conclusion}
The findings from this study highlight the potential utility of LLMs in generating clinically relevant descriptions of DFUs. With its strong performance in accurately capturing clinical features, GPT-4o demonstrated promise as an assistive tool in clinical settings. By providing reliable and detailed descriptions of DFU images, such LLMs can help streamline the documentation process, reduce clinicians' workload, and improve the consistency of patient records in EMR systems, facilitating effective triage systems for early detection and treatment.

However, the performance variability observed among the other models highlights the need for ongoing refinement and specialization of LLM tools in healthcare. While promising, open-source models like Qwen-VL still require significant optimization to match the performance of proprietary models like GPT-4o. The application of LLM in DFU management can also lead to the development of integrated telemedicine systems, where remote monitoring and assessment of DFUs become more efficient and scalable.

Future work should focus on validating these findings with larger datasets and refining the evaluation process to ensure model outputs are clinically accurate, comprehensive, and diagnostically useful in various settings. As LLMs evolve, their role in clinical practice will likely grow, offering clinicians powerful tools to enhance patient care, particularly for chronic conditions like DFUs requiring ongoing monitoring.

\subsubsection{Acknowledgements} 
The authors sincerely thank Dr. Karim Manji and Dr. John Toole from the Zivot Limb Preservation Centre, Calgary, for their invaluable contributions to this study. Their expertise and thorough evaluations of the generated text were critical in assessing the clinical relevance and accuracy of the language models tested. We greatly appreciate their time and dedication to this project.

%
%
%
%

\bibliographystyle{splncs04}
\bibliography{egbib}

\end{document}